\documentclass[conference]{IEEEtran}
\IEEEoverridecommandlockouts
\usepackage{cite}
\usepackage{amsmath,amssymb,amsfonts}
\usepackage{algorithmic}
\usepackage[ruled]{algorithm2e} 
\usepackage{graphicx}
\usepackage{textcomp}
\usepackage{xcolor}
\usepackage{flushend}     
\usepackage[caption = false]{subfig}

\def\BibTeX{{\rm B\kern-.05em{\sc i\kern-.025em b}\kern-.08em
    T\kern-.1667em\lower.7ex\hbox{E}\kern-.125emX}}
    
\makeatletter
\newcommand{\linebreakand}{%
  \end{@IEEEauthorhalign}
  \hfill\mbox{}\par
  \mbox{}\hfill\begin{@IEEEauthorhalign}
}
\makeatother

\begin{document}

\title{Foveal-pit inspired filtering of DVS spike response\\
{\footnotesize \textsuperscript{}}
\thanks{}
}

\author{\IEEEauthorblockN{1\textsuperscript{st} Shriya T.P. Gupta}
\IEEEauthorblockA{\textit{Microsoft Research and Development Center} \\
\textit{Microsoft Corporation Pvt. Ltd.}\\
Bangalore, India \\
shriyatp99@gmail.com}
\and
\IEEEauthorblockN{2\textsuperscript{nd} Pablo Linares-Serrano}
\IEEEauthorblockA{\textit{Instituto de Microelectrónica de Sevilla } \\
\textit{IMSE-CNM (CSIC and Universidad de Sevilla)}\\
Sevilla, Spain \\
pablolinareserrano@gmail.com}
\linebreakand
\IEEEauthorblockN{3\textsuperscript{rd} Basabdatta Sen Bhattacharya}
\IEEEauthorblockA{\textit{Department of Computer Science} \\
\textit{BITS Pilani Goa Campus}\\
Goa, India \\
basabdattab@goa.bits-pilani.ac.in}
\and
\IEEEauthorblockN{4\textsuperscript{th} Teresa Serrano-Gotarredona}
\IEEEauthorblockA{\textit{Instituto de Microelectrónica de Sevilla } \\
\textit{IMSE-CNM (CSIC and Universidad de Sevilla)}\\
Sevilla, Spain \\
terese@imse-cnm.csic.es}
}

\maketitle

\begin{abstract}
In this paper, we present results of processing Dynamic Vision Sensor (DVS) recordings of visual patterns with a retinal model based on foveal-pit inspired Difference of Gaussian (DoG) filters. A DVS sensor was stimulated with varying number of vertical white and black bars of different spatial frequencies moving horizontally at a constant velocity. The output spikes generated by the DVS sensor were applied as input to a set of DoG filters inspired by the receptive field structure of the primate visual pathway. In particular, these filters mimic the receptive fields of the midget and parasol ganglion cells (spiking neurons of the retina) that sub-serve the photo-receptors of the foveal-pit. The features extracted with the foveal-pit model are used for further classification using a spiking convolutional neural network trained with a backpropagation variant adapted for spiking neural networks.
\end{abstract}

\begin{IEEEkeywords}
dynamic vision sensor, neural filtering, spiking neural network, classification, difference of gaussian, convolution, foveal-pit
\end{IEEEkeywords}

\section{Introduction}
Recent advances in deep learning~\cite{bengio2007scaling, lecun2015deep} have led to state-of-the-art performance for varied classification tasks in natural language processing, computer vision and speech recognition. Traditional Artificial Neural Networks (ANN) use idealized computing units which have a differentiable, non-linear activation function allowing stacking of such neurons in multiple trainable layers. The existence of derivatives makes it possible to carry out large scale training of these architectures with gradient based optimization methods~\cite{lecun1998gradient} using high computing resources like Graphic Processing Units (GPU). However, this prevents the use of such deep learning models for essential real-life applications like mobile devices and autonomous systems that have limited compute power.

Spiking Neural Networks (SNN) have been proposed as an energy-efficient alternative to ANNs as they simulate the event-based information processing of the brain~\cite{gerstner2002spiking}. These bio-inspired SNNs follow an asynchronous method of event processing using spiking neurons. The internal state of a spiking neuron is updated when it receives an action potential and consequently an output spike is fired when the membrane voltage crosses a pre-defined threshold. Further, improvements in neuromorphic engineering allow the implementation of SNNs on neuromorphic hardware platforms~\cite{young2019review} that lead to a much higher efficiency in terms of power and speed compared to conventional GPU based computing systems.

Although SNNs are considered as the third generation of  neural networks holding the potential for sparse and low-power computation, their classification performance is considerably lower than those of ANNs. This can be attributed to the fact that gradient optimization techniques like the backpropagation algorithm can't be implemented in SNNs due to the discrete nature of spiking neurons. A common technique for training SNN models is the Hebbian learning inspired Spike Timing Dependent Plasticity (STDP) that is used in several state-of-the-art approaches~\cite{kheradpisheh2018stdp, diehl2015fast}. Other works like~\cite{neftci2017event, mostafa2017supervised} have adapted the gradient descent algorithm for SNNs using a differentiable approximation of spiking neurons. Our approach also employs a similar modified backpropagation algorithm proposed by Hunsberger et al.~\cite{hunsberger2016training} that is implemented in the Nengo-DL library~\cite{bekolay2014nengo}.

As shown by Camunas-Mesa et al.~\cite{camunas2011event}, the efficiency gain of SNNs from event-based processing can be further improved through the use of inputs from event-based sensors like a neuromorphic Dynamic Vision Sensor (DVS)~\cite{serrano2013128}. Event driven sensors represent the information dynamically by asynchronously transmitting the address event of each pixel and hence avoid processing redundant data. However, the classification accuracy drops drastically when using real sensory data from a physical spiking silicon retina, since the spike events are no longer Poissonian~\cite{orchard2015converting}.
\begin{figure*}[htbp] 
\centering
      \includegraphics[scale=0.45]{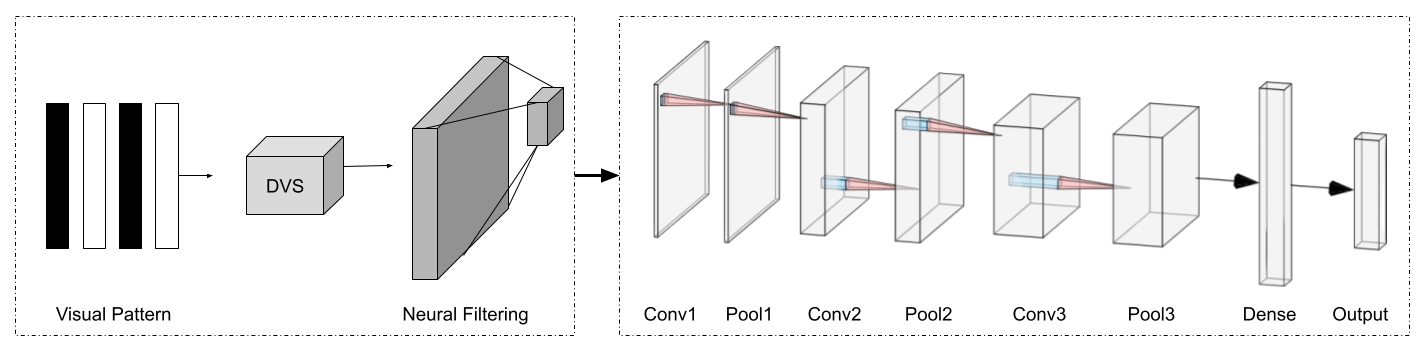}
      \caption{The two-stage architecture of the proposed DVS based spiking convolutional neural network. In the first stage, the DVS is stimulated with a pattern of vertical white and black bars and the generated spike responses are processed using foveal-pit inspired DoG filters. The second stage is composed of the convolutional and pooling layers which are used to classify the features extracted from the first stage.}
      \label{CNNnet}
\end{figure*}

In our previous work~\cite{gupta2020implementing}, we had demonstrated the effect of foveal-pit inspired filtering for synthetically generated datasets like MNIST~\cite{lecun1998gradient} and Caltech~\cite{fei2004learning}. In this work, we present the results of applying similar neural filtering to data generated by the DVS. In our proposed model, we process DVS outputs using bio-inspired filters that simulate receptive fields of the midget and parasol ganglion cells of the primate retina. The DVS is stimulated with vertical black and white bars having a constant displacement of 2 pixels from frame to frame. The foveal-pit informed Difference of Gaussian (DoG) filters are applied to the DVS recordings in order to capture the most perceptually important information from the input data. The use of DoG functions to model retinal filters was originally proposed by Rullen et al.~\cite{rullen2001rate} and the receptive fields of the foveal-pit are implemented as in Bhattacharya et al.~\cite{bhattacharya2010biologically}. 

The processed features are then used to perform the classification using a Spiking Convolutional Neural Network (SCNN). The SCNN architecture is inspired by two previous works viz. Diehl et al.~\cite{diehl2015fast} and Kheradpisheh et al.~\cite{kheradpisheh2018stdp}, while the model is implemented as in Gupta et al.~\cite{gupta2020implementing}. Each input is presented to the network for a total duration of 60 timesteps and the predictions are assigned based on the voltages measured from the output neurons. The empirical results demonstrate that the application of neural filtering to DVS recordings leads to an improvement of 35\% in classification accuracy compared to the unfiltered DVS spike responses. Out of the filtered scenarios, the highest performance of 100\% is achieved using the off-center parasol ganglion cells. 

The rest of the paper is organized as follows: Section II describes the architecture of the proposed model including the response generation and filtering, Section III provides the results of the experiments and Section IV contains the conclusion and future directions.

\section{Methodology}
The overall architecture of our model consists of two main stages: the first stage is made up of the DVS response generation and neural filtering of output spikes; the second stage consists of performing classification using the SCNN. The proposed model is shown in Fig.~\ref{CNNnet} and each of the individual stages are covered in detail in the following sub-sections.

\subsection{Dynamic Vision Sensor Responses}
\begin{figure}[hbtp] 
\centering
      \includegraphics[scale=0.45]{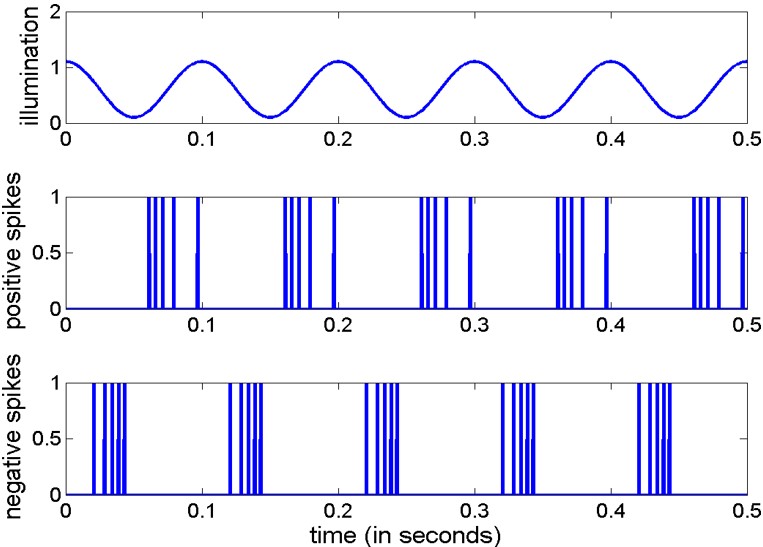}
      \caption{Illustration of the DVS spike response generated using a sinusoidal input stimulus. The first row represents the pixel's illumination over time. The second row depicts the positive spikes whereas the last row represents the negative spikes corresponding to a decrease in illumination~\cite{sen2017spiking}.}
      \label{dvs_spike}
\end{figure}

We have used a $128\times128$ sized neuromorphic DVS developed by Serrano-Gotarredona et al.~\cite{serrano2013128} to capture non-synthetic visual data. Each pixel of the DVS processes the input continuously and emits a spike based on the variation in the illumination impinging upon it~\cite{sen2017spiking}. A sample illustration is provided in Fig.~\ref{dvs_spike} using a sinusoidal input stimulus having a frequency of 10Hz. The first row depicts the pixel's illumination over time whereas the remaining two rows capture the emission of spikes over the same duration corresponding to changes in illumination. An increase in illumination leads to a positive spike whereas a decrease in illumination leads to a negative spike as seen in the last row of Fig.~\ref{dvs_spike}.

\begin{figure}[hbtp] 
\centering
      \includegraphics[scale=0.15]{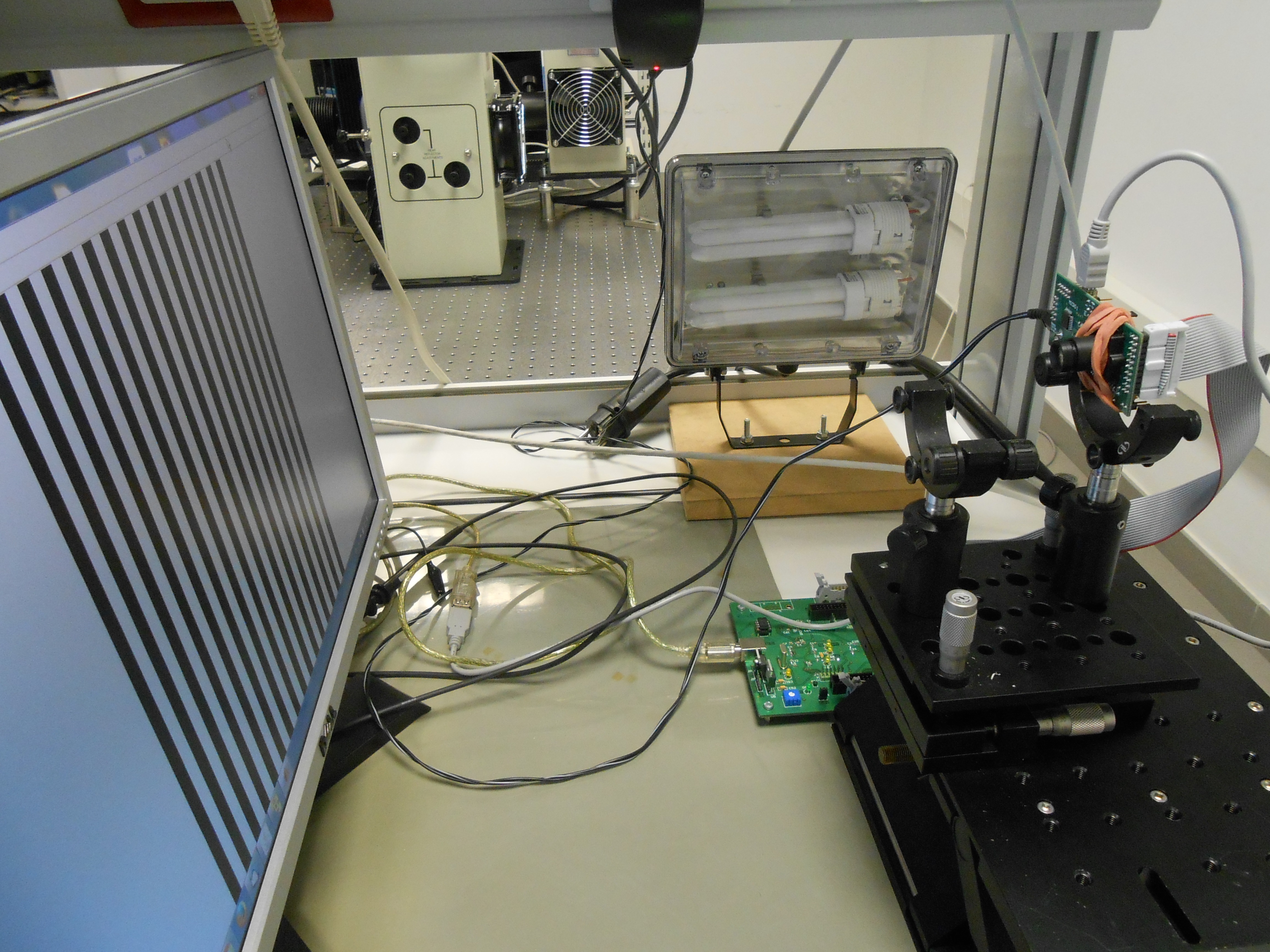}
      \caption{The DVS setup to record spike responses when presented with a simple pattern moving across a computer screen as visual stimulus.}
      \label{dvs}
\end{figure}

For our experiments, the DVS was placed in front of a monitor displaying a pattern of equally wide black and white vertical bars as shown in Fig.~\ref{dvs}. The bars were moved horizontally across the screen such that a displacement of 2 pixels is applied from frame to frame. The number of bars were varied from 2, 4, 8, 16, 32, 64 to 128 and these $K=7$ categories correspond to the final labels for our multiclass classification problem. The events generated by the DVS were captured in the Address Event Representation (AER) format using the jAER software~\cite{jaer}.

\subsection{Retina-inspired filtering}
The DVS recordings generated from the first stage of our model are passed to a set of neural filters simulating the primate visual system. As proposed by Kheradpisheh et al.~\cite{kheradpisheh2018stdp}, we have used DoG functions to implement these biologically inspired filters sub-serving the retinal foveal pit. The foveal pit is a circular region of 200 $\mu m$ diameter that lies at the center of the foveola. This region also has the highest visual acuity in the primate retina and is most accessible to incoming light. The fovea is sub-served by a ganglion cell layer composed of midget and parasol cells. The retinal ganglion cells are the only spiking neurons of the primate visual system and their axons transmit the received information from the retina to other parts of the brain. 
\begin{algorithm}
 \label{algo}
 \caption{Algorithm for filtering the DVS spike response using foveal-pit inspired DoG functions.}
 \begin{algorithmic}[1]
    \STATE kernel = dog\_func(mat\_dim, cent\_dev, circ\_shift)
    
    \STATE[ma, na] = size(input)
    \STATE[mb, nb] = size(kernel)
    \STATE filt\_out = zeros(ma, na)
    \STATE r1 = ceil(mb/2)
    \STATE s1 = ceil(nb/2)

    \FOR{i = 1 to i = ma}
       \FOR{j =1 to j = na}
           \STATE i1 = max(0, i-r1);                
           \FOR{r = max(1, r1-i+1) to r = mb}        
             \STATE i1 = i1 + 1
             \STATE j1 = max(0, j-s1)
                   \FOR{s = max(1, s1-j+1) to s = nb}
                    \STATE j1 = j1 + 1
                         \STATE filt\_out(i, j) += kernel(r, s) * input(i1, j1)
                     \ENDFOR
                \ENDFOR
           \ENDFOR
     \ENDFOR
    \RETURN filt\_out
 \end{algorithmic} 
\end{algorithm}

The midget and parasol ganglion cells have two types of centre surround receptive fields --- on-centre-off-surround and off-centre-on-surround. We have modelled these receptive fields using DoG functions as specified in Bhattacharya et al.~\cite{bhattacharya2010biologically}. The off-center midget cells have a matrix size of $5\times5$ with standard deviation of 0.8 whereas the on-center midget cells are of size $11\times11$ with standard deviation of 1.04. Similarly, the off-center parasol cells have a size of $61\times61$ with a standard deviation of 8 while the on-center parasol cells are of size $243\times243$ with a standard deviation of 10.4. These DoG functions are then applied to the DVS spike responses using Algorithm~\ref{algo}. 

\subsection{Convolutional Network Architecture}
The asynchronous DVS recordings generated from the previous stage are split into individual frames for training our frame-based classifier composed of convolutional layers. This modified dataset is created following the procedure of Stromatias et al.~\cite{stromatias2017event} to produce an analog vector representation. The SCNN architecture used in our work consists of three convolutional and pooling layers which are made up of Leaky Integrate and Fire (LIF) neurons. 
\begin{table}[!thpb]
\renewcommand{\arraystretch}{1.0}
\caption{Dimensions of the SCNN layers.}
\label{dimension}
\centering
 \begin{tabular}{|c|c|c|c|} 
 \hline
Layer & No. of filters & Input size & Kernel size\\[0.5ex]
\hline
Conv1 & 8 &(128, 128) & 3 \\
Pool1 & - &(128, 128)& 2 \\
Conv2 & 16 &(64, 64)& 3 \\
Pool2 & - &(64, 64) & 2 \\
Conv3 & 32 &(32, 32) & 3 \\
Pool3 & - &(32, 32) & 2 \\
Flatten & - &(16, 16) &  - \\
Dense &  -  &(1, 8192) & - \\
Outputs & - & (1, 7) & -\\
 \hline
\end{tabular}
\end{table}

Traditional deep learning architectures use sigmoid neurons which are differentiable non-linearities, whereas the spiking neurons used in SCNNs are non-differentiable. Hence, we use a differentiable approximation of the spiking neurons during training and the actual spiking neurons during inference as proposed by Hunsberger et al.~\cite{hunsberger2016training}. Since we use a rate-based approximation during training, the model is run only for a single timestep whereas during testing with the spiking neurons, the model is run for 60 timesteps to collect the cumulative spike output over time.

The convolution is carried out on the $128\times128$ input arrays using filters of size $3\times3$. The first, second and third convolutional layers of the SCNN are made up of $2^3$, $2^4$ and $2^5$ filters respectively, followed by a pooling operation after each convolution. The synaptic connections between the neurons of these layers are modelled as the trainable weights of the network which are optimized by minimizing the loss function of the overall SCNN. The exact dimensions of the individual layers are provided in Table~\ref{dimension}.

\subsection{Training and Inference}

For our multiclass classification problem with $K=7$ categories, we convert the outputs of the last pooling layer into a 1-D vector using a flatten operation. This is followed by a dense layer with all-to-all connectivity having K neurons which generates a $K\times1$ output vector. A softmax classifier is used to transform these output values into a set of K probabilities:
\begin{equation}
\label{eq:f2}
\mathbf{Y}(X, W) =  \frac{ e^{w_i x_i} } {\sum_{j=1}^{K} e^{w_j x_j} } \forall  i = 1,\cdots,K
\end{equation}
where $x_i \in X$  and $w_i \in W$ are the inputs and weights of the dense layer respectively, and $\mathbf{Y}$ is the prediction probabilities that sum to 1. The Negative Log Likelihood (NLL) loss for the overall network is computed using the one-hot encoded output labels $\mathbf{L}$ and the softmaxed probabilities $\mathbf{Y}$ with NLL defined as:
\begin{equation}
\label{eq:f3}
\mathbf{O}(X, W) = - \frac{1}{M} \sum_{i}^{M} \sum_{j}^{K} F_i(j) * \log (\mathbf{Y}(X, W))
\end{equation}
where M is the mini-batch size, $F_i(j) = 1$ when $j = L_i$ and zero otherwise. The SCNN is trained end-to-end using a spiking approximation of the backpropogation algorithm adapted for SNNs. This is done by minimizing the NLL loss using the procedure described in Gupta et al.~\cite{gupta2020implementing} with a duration of 3 epochs and a mini-batch size of 20.

For the inference stage, we pass the input images from the testing corpora and measure the voltages (mV) of the output layer neurons. These values are generated using the probe function of the Nengo-DL library and represent the progressively increasing membrane potentials. Thus, the neuron having the highest voltage over a 60 ms simulation time period is assigned as the predicted class for that epoch.

\section{Experimental Methods and Results}
The filtered spikes responses from the 128$\times$128 sized DVS sensor was split into individual frames for each recording to be passed as input to the subsequent convolutional network. This resulted in a total of 3552 images for the unfiltered scenario and a collection of 3503 images for the filtered recordings. In each case, the images were then partitioned in the ratio of 9:1 to create the corresponding training and testing corpora. 

Generation of the DVS spikes responses along with the filtering was implemented entirely in Matlab, while the SCNN and its various layers were coded in Python using the Nengo-DL library~\cite{bekolay2014nengo}. The generated .mat files of the dataset were loaded into the Python network using the Scipy library~\cite{2020SciPy-NMeth} and the experiments were carried out on the GPU accessed via Google Colaboratory~\cite{colab}. 

\subsection{Quantitative Effects of Filtering}
\label{sec:3a}
To assess the effects of incorporating the neural filtering on DVS recordings, we ran two experiments with the SCNN for a total duration of 60 timesteps. The empirical results are summarized in Table~\ref{results}. For the first scenario of using unfiltered DVS frames, the model achieves an accuracy of 65\% which is significantly lower than the values in the remaining rows that correspond to the filtered DVS inputs. This demonstrates that introducing the foveal-pit inspired neural filtering into our retinal model leads to a considerable improvement of 35\% even for simplistic visual patterns such as those in our dataset. 
\begin{table}[!thpb]
\renewcommand{\arraystretch}{1.0}
\caption{Accuracies (\%) for the frame-based DVS input}
\label{results}
\centering
 \begin{tabular}{|c|c|c|c|} 
 \hline
Scenario & Cell - Type  &  CircShift  & Accuracy \\[0.5ex]
\hline
Unfiltered & - & - & 65.0 \% \\
\hline
 & off-center midget &  & 77.5 \% \\
Filtered & on-center midget & 0 & 85.0 \% \\
 & off-center parasol &  & 92.5 \% \\
 & on-center parasol &  & 87.5 \% \\
 \hline
 & off-center midget &  & 77.5 \% \\
Filtered & on-center midget &  1 & 85.0 \% \\
 & off-center parasol &  & 100.0 \% \\
 & on-center parasol &  & 85.0 \% \\
\hline
\end{tabular}
\end{table}

Amongst the filtered outputs, the parasol ganglion cells have a comparatively higher increase in accuracy compared to the midget cells. Since the parasol cells have larger dimensions and capture the overall background information, they lead to a significant improvement in classification of distinct patterns without intricate details. Thus, the parasol cells lead to a larger performance gain achieving a highest accuracy in the shifted case, as our dataset is composed of only vertical black and white bars. On the other hand, midget cells have smaller dimensions which allows them to capture only the finer details of an image and hence they contribute lesser to the overall increase in classification accuracy. 

From Table~\ref{results}, we can also observe that the variations in accuracy for filtering with different ganglion cell types is almost comparable for both the cases of with and without any circular-shift; circular-shift refers to the case where the DoG filters are circular shifted to `wrap' on the raster at all four edges. This reduces artifacts due to edges. The alternative scenario is zero padding at all four edges. The circular-shift value in the Table~\ref{results} is set to 1 indicating all cases where filtering was performed using circular shift at edges, and is set to 0 otherwise.

\subsection{Qualitative Effects of Filtering}
\begin{figure}[thpb] 
\centering
\subfloat[Unfiltered scenario]{\includegraphics[width = 2in]{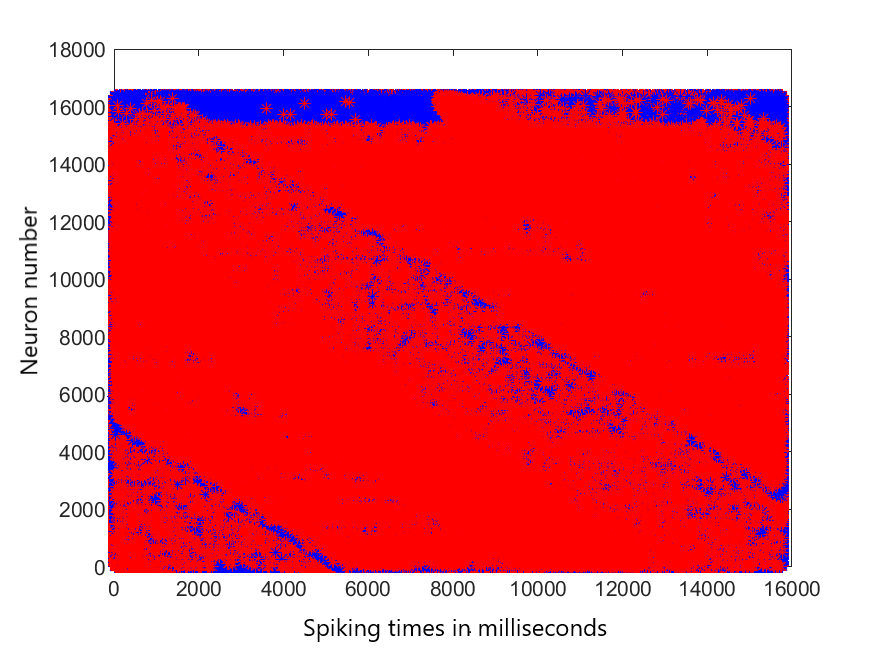}}
\\
\subfloat[Filtered: Unshifted midget]{\includegraphics[width = 2in]{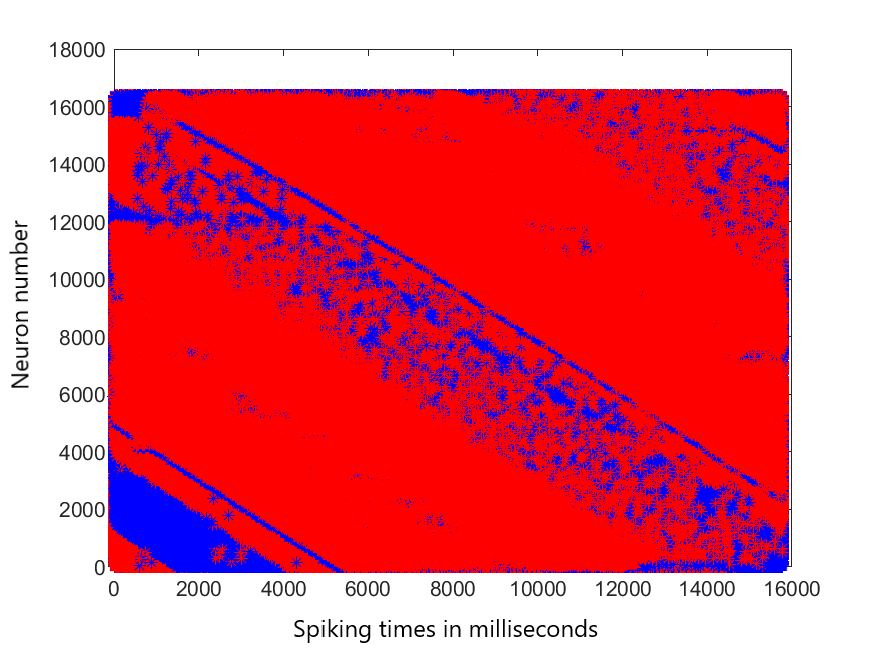}}
\\
\subfloat[Filtered: Unshifted parasol]{\includegraphics[width = 2in]{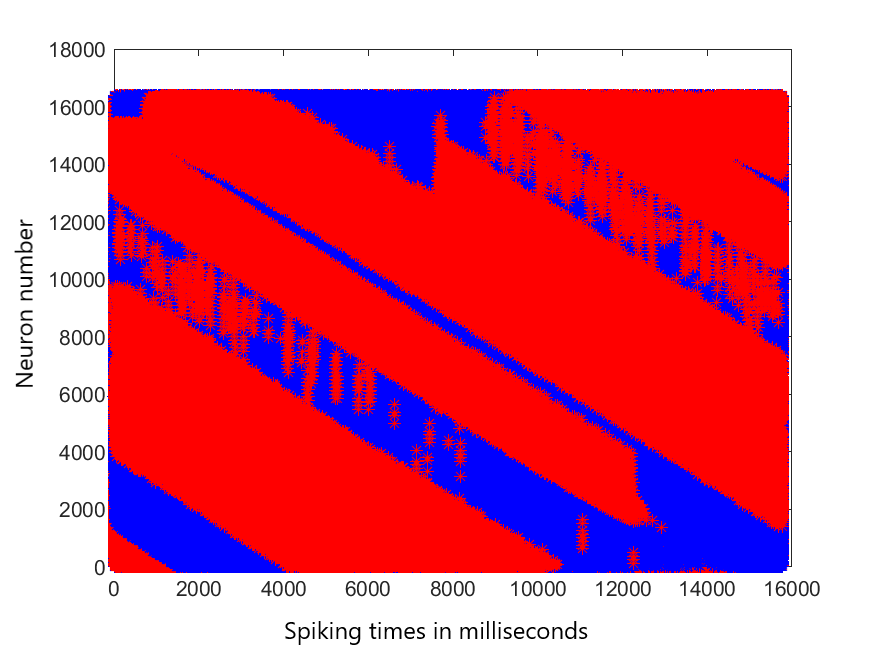}}
\\
\subfloat[Filtered: Circular-shifted midget]{\includegraphics[width = 2in]{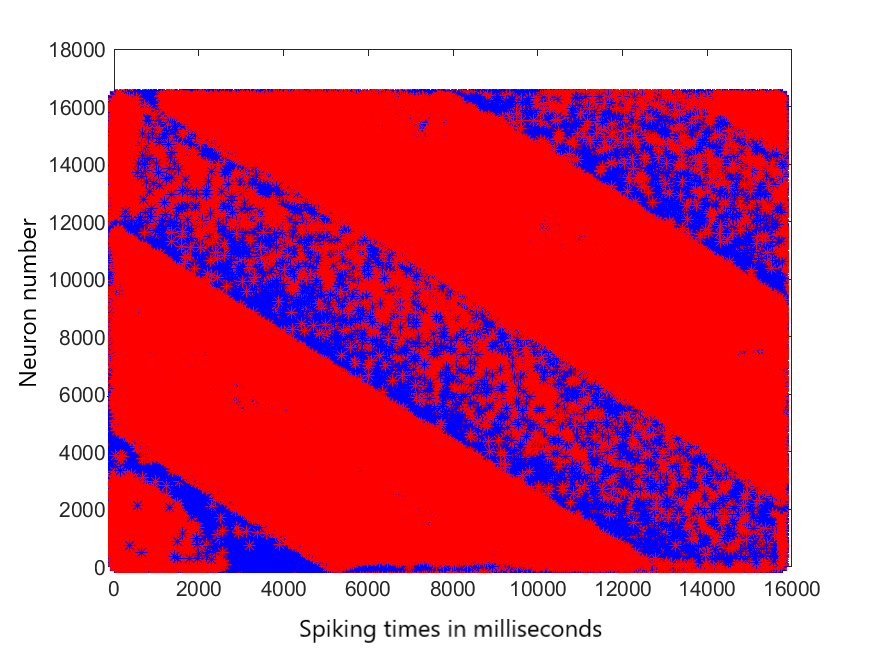}}
\\
\subfloat[Filtered: Circular-shifted parasol]{\includegraphics[width = 2in]{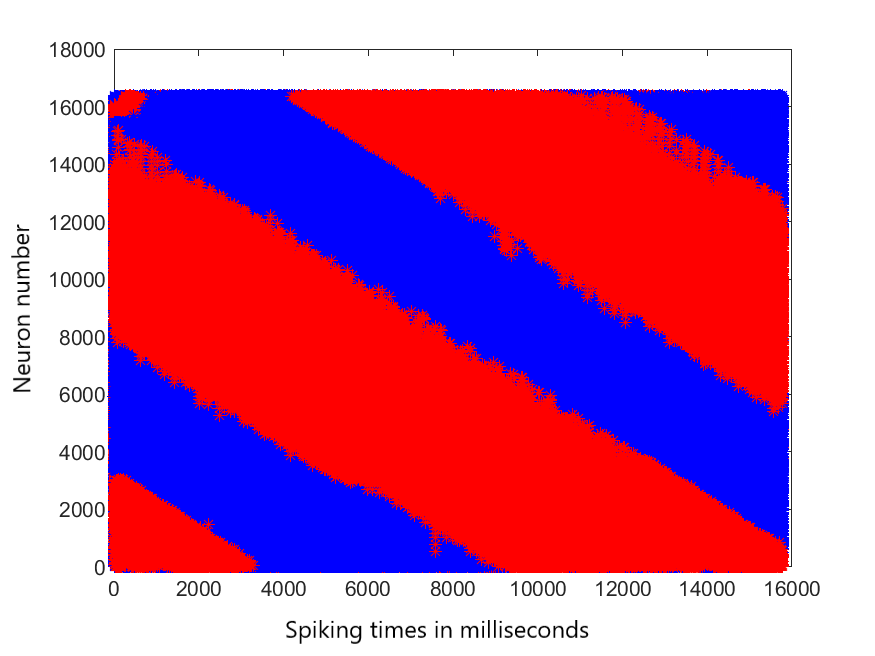}}
\caption{Raster plot of the (a) unfiltered DVS spike response and filtered DVS response using the off-center (b) unshifted midget cell (c) unshifted parasol cell (d) circular-shifted midget cell (e) circular-shifted parasol cell.}
\label{rasters}
\end{figure}

For analysing the qualitative effects of neural filtering, we generated raster plots using the analog vector representation of DVS responses as shown in Fig.~\ref{rasters}. The neuron numbers range from a value of 0 to 16384 as they represent the pixels of the 128$\times$128 sized electronic retina used in our experiments. The blue markers depict a positive event corresponding to an increase in illumination as the moving edges go from black to white. On the contrary, the red markers represent a negative event and indicate a decrease in illumination as the edges go from white to black. 

Figure~\ref{rasters}(a) illustrates the unfiltered scenario which has the least distinction between positive and negative events of the input stimulus. This lack of differentiation between black and white bars of the visual pattern also leads to a drop in the classification accuracy which was previously observed in Table~\ref{results}. Further, in the filtered raster plots of Fig.~\ref{rasters}, we note that all the ganglion cell filters capture edges more effectively compared to the unfiltered case as there is a clear distinction in the positive and negative events of the filtered raster plots. This improved distinction also leads to a higher classification performance as seen in Section~\ref{sec:3a}. 

Amongst the outputs of the filtered DVS responses, Fig.~\ref{rasters}(b) and Fig.~\ref{rasters}(c) represent the cases having a circular-shift value of 0 while Fig.~\ref{rasters}(d) and Fig.~\ref{rasters}(e) correspond to a circular-shift value of 1. The variation in filtered responses are almost similar for these two cases as seen in Table~\ref{results}. However, the result obtained using the unshifted parasol cell in Fig.~\ref{rasters}(c) has lesser clarity between the positive and negative events and hence contains less distinguishable edges than in Fig.~\ref{rasters}(e) using the circular-shifted filter. Thus, the best classification accuracy of 100\% in Section~\ref{sec:3a} was achieved using the circular-shifted parasol filter as it captures the changes in illumination more effectively.

Additionally, we observe that the plots generated using midget ganglion cells leads to a noisier qualitative output than those using the parasol ganglion cells. This is because the midget cells have smaller dimensions and are able to pick up only the finer details of an image. But since the simplistic visual patterns used in our experiments lack any intricate details, the midget cells contribute lesser to the overall improvement in the classification performance. Thus, the highest accuracies in Table~\ref{results} are obtained using the parasol cells as they capture the larger and more significant information contained in the input data.

\section{Conclusion and Future Work}
In this paper, we have presented a novel method for processing the DVS spike responses of a visual pattern with foveal-pit inspired DoG filters that simulate the primate retinal system. The pattern was composed of varying number of vertical white and black bars of different spatial frequencies moving at a fixed velocity. The outputs from the sensor are applied as input to the bio-inspired neural filters that model the receptive field structure of midget and parasol ganglion cells of the foveal-pit. These processed features are passed as input to our spiking convolutional neural network architecture which classifies the frame-based version of the filtered responses into seven corresponding categories. The SCNN is composed of convolutional and pooling layers and is trained with a modified backpropogation algorithm using a differentiable approximation of spiking neurons~\cite{hunsberger2016training}.

The proposed model demonstrates the effect of applying neural filtering to real DVS data generated from a neuromorphic vision sensor. This builds upon our previous work~\cite{gupta2020implementing} that depicted the results of foveal-pit inspired filtering for synthetically generated datasets like MNIST~\cite{lecun1998gradient} and Caltech~\cite{fei2004learning}. Our model achieves a promising performance of 92.5\% using the unshifted off-center parasol ganglion cell and an accuracy of 100\% in the circular-shifted scenario, which is an improvement of 35\% over the classification using unfiltered DVS responses. The empirical results indicate the importance of the foveal-pit inspired neural filtering in redundancy reduction of the DVS inputs and in discarding irrelevant background information. 

For our proposed network, the asynchronous DVS recordings generated from the first stage of the model were converted to an analog vector representation for training the frame-based classifier composed of convolution layers. As future work, we plan to adapt our spiking convolutional network architecture to directly process event-based data and evaluate the effects of the bio-inspired neural filtering on continuous outputs of a neuromorphic DVS. Also, the dataset used in this work is limited in terms of variation in the inputs as well as the size of the training and testing corpora. Hence, we would like to further verify the effect of the DoG filters on DVS spike responses of larger and more complex datasets. 

\section{Acknowledgements}

This work was funded by the EU grant PCI2019-111826-2 “APROVIS3D”, by Spanish grant from the Ministry of Science and Innovation PID2019-105556GB-C31 “NANOMIND” (with support from the European Regional Development Fund) and by the CSIC 2018-50E008 AVE project. BSB is supported by the Science and Engineering Research Board (SERB) of India Fund CRG/2019/003534.

\bibliographystyle{IEEEtran}
\bibliography{IEEEabrv, ref}

\end{document}